\def\year{2019}\relax
\documentclass[letterpaper]{article} 
\usepackage{aaai19}  
\usepackage{times}  
\usepackage{helvet}  
\usepackage{courier}  
\usepackage{url}  
\usepackage{graphicx}  
\graphicspath{ {./images/} }
\usepackage{hyperref}
\usepackage{balance}
\usepackage{bbm}

\usepackage{caption}
\usepackage{amssymb}

\def\R{{\mathbb{R}}}

\usepackage{booktabs} 
\usepackage{enumitem}

\usepackage{tabu}
\usepackage{adjustbox}
\usepackage{multirow}
\usepackage{comment}
\usepackage{amsmath}
\begin{document}
%
\title{A Memory-Network Based Solution for Multivariate Time-Series Forecasting}
\author{Yen-Yu Chang$^1$, Fan-Yun Sun$^1$, Yueh-Hua Wu$^{1,2}$, Shou-De Lin$^1$ \\ $^1$National Taiwan University, $^2$Riken-AIP \\ \{b03901138,b04902045\}@ntu.edu.tw, \{d06922005, sdlin\}@csie.ntu.edu.tw}
\maketitle
\begin{abstract}

 Multivariate time series forecasting is extensively studied throughout the years with ubiquitous applications in areas such as finance, traffic, environment, etc. Still, concerns have been raised on traditional methods for incapable of modeling complex patterns or dependencies lying in real word data. To address such concerns, various deep learning models, mainly Recurrent Neural Network (RNN) based methods, are proposed. Nevertheless, capturing extremely long-term patterns while effectively incorporating information from other variables remains a challenge for time-series forecasting. Furthermore, lack-of-explainability remains one serious drawback for deep neural network models. Inspired by Memory Network proposed for solving the question-answering task, we propose a deep learning based model named Memory Time-series network (MTNet) for time series forecasting. MTNet consists of a large memory component, three separate encoders, and an autoregressive component to train jointly. 
 Addtionally, the attention mechanism designed enable MTNet to be highly interpretable. We can easily tell which part of the historic data is referenced the most. 
\end{abstract}
\section{Introduction}

Multivariate time series forecasting has been widely applied in many areas such as financial market prediction \cite{wu2013dynamic}, weather forecasting \cite{chakraborty2012fine}, complex dynamical system analysis \cite{liu2015regularized}, and the analysis for Internet-of-Things data where multiple sensors are deployed to collect useful real-time information \cite{yang2015deep}. Given multiple time series where some or all of them are to certain extend correlated, how to discover and leverage the dynamics and dependencies among them while making predictions in reasonable time has become a major challenge. 

There have been many solutions proposed for time series forecasting such as vector autoregression (VAR) \cite{box2015time,lutkepohl2005new} autoregressive moving average (ARMA) model \cite{whitle1951hypothesis}, autoregressive integrated moving average (ARIMA) model  \cite{box2015time}, and regression based methods such as linear support vector regression (SVR) \cite{cao2003support,kim2003financial}. 
These models usually assume certain distribution or function form of time series and may not be able to capture the complex underlying nonlinear relationships. Other models such as Gaussian Process \cite{roberts2013gaussian} entails high computational cost to handle data of larger size.

Deep neural networks, especially recurrent neural networks (RNN) based models, have been proposed for time series forecasting recently due to their success in other related domains. For example, encoder-decoder networks \cite{kalchbrenner2013recurrent,cho2014properties,sutskever2014sequence} have its major success in machine translation and other natural language processing (NLP) tasks. Such RNN-based methods and their variations, including the early work using naive RNN models \cite{connor1992recurrent}, hybrid models combining the use of ARIMA and Multilayer Perceptron (MLP) \cite{jain2007hybrid,zhang1998forecasting,zhang2003time}, or the recent combination of vanilla RNN and Dynamic Boltzmann Machines \cite{dasgupta2017nonlinear}, have been shown to outperform the non deep learning models in time-series forecasting. Despite the success of RNN based models, they still may fail on tasks that require long-term information, mainly due to the effect of gradient vanishing.
As shown in previous work \cite{cho2014properties}, encoder-decoder networks deteriorate rapidly as the length of the input sentences increases. As a remedy to the vanishing gradient problem, researchers have proposed the 
attention-based encoder decoder network\cite{bahdanau2014neural}. Attention mechanism has gained increasing popularity due to its ability in enabling recurrent neural networks to select parts of hidden states across time steps.

Long- and Short-term Time-series Network (LSTNet) is considered as a reliable solution for multivariate time series forecasting  \cite{lai2017modeling}. It leverages both the convolutional layer and the recurrent layer to
capture complex long-term dependencies. A novel recurrent structure, namely Recurrent-skip, is
designed for capturing long-term dependence and making the optimization easier as it utilizes the periodic property of the input time series signals. LSTNet also incorporates a traditional autoregressive linear model in parallel to the non-linear neural network to improve the robustness of the model. One major downside of it is that Recurrent-skip layer requires a predefined hyperparameter $p$, which is unfavorable in time series whose period length is dynamic over time. To alleviate the issue, an alternative model replacing Recurrent-skip layer with Temporal attention layer has been proposed.  

The dual-stage attention based recurrent neural network (DA-RNN) \cite{qin2017dual} is considered as the state-of-the-art solution in time series prediction. The model is divided into two stages.
In the first stage, an attention mechanism is proposed to adaptively extract the relevant driving series at each time step by referring to the previous encoder hidden state. In the second stage, a temporal attention mechanism is used to select relevant encoder hidden states across all time steps.
However, DA-RNN does not consider the “spatial” correlations among different components of exogenous data. 
More importantly, in the second stage DA-RNN conducts point-wise attention which might not be suitable for capturing continuous periodical patterns, as will be discussed through our experiments later. 

To address the aforementioned concerns, we propose to exploit the idea of Memory network \cite{DBLP:journals/corr/WestonCB14} to handle the time-series forecasting task. Memory network was first proposed to use in the context of question answering, combining inference components with a long-term memory component and learn how to use these jointly. End-to-end memory network \cite{sukhbaatar2015end} is later introduced as a refinement to the previous work. It achieved competitive results on question answering compared to the previous work (\cite{DBLP:journals/corr/WestonCB14}), while requiring significantly less supervision.

Inspired by the concept of an end-to-end Memory Network \cite{sukhbaatar2015end}, we proposed a time series forecasting model that consists of a memory component, three different embedding feature maps generated by three independent encoders, and an autoregressive component. The memory component is used to store the long-term historical data, while encoders are used to convert the input data and memory data to their feature representations. After calculating the similarity between the input data and the data stored in the memory component, we derived attentional weights across all chunks of memory data. As in LSTNet, we also incorporate a traditional autoregressive linear model in parallel to the non-linear neural network to improve the robustness of our model. 
Compared to the existing non-memory based RNN models and their variations, the proposed memory component and attention mechanism are more effective in capturing very long-term dependencies as well as periodic patterns in time series signals. In particular, our attention differs from the temporal attention used in both LSTNet and DA-RNN, which only allows the attention to a particular \textit{timestamps} in the past. The proposed model is capable of attending to \textit{a period of time}, which can be considered as a chunk of memory in our framework.

Furthermore, the proposed model is naturally extendable to both univariate and multivariate settings, unlike DARNN which is more suitable for univariate time series forecasting. The first-stage attention mechanism of DA-RNN extracts relevant driving series at each timestep. In multivariate settings, every single target variable will require their own attention layer since different variable should be able to focus on different driving series at each timestep. That makes DA-RNN computationally expensive for multivariate time series forecasting.
In MTNet, we simply utilize convolutional filters to incorporate correlations among different dimension features.

\begin{figure*}[h]
\centering
\resizebox{\textwidth}{!}{
\includegraphics[width=\textwidth]{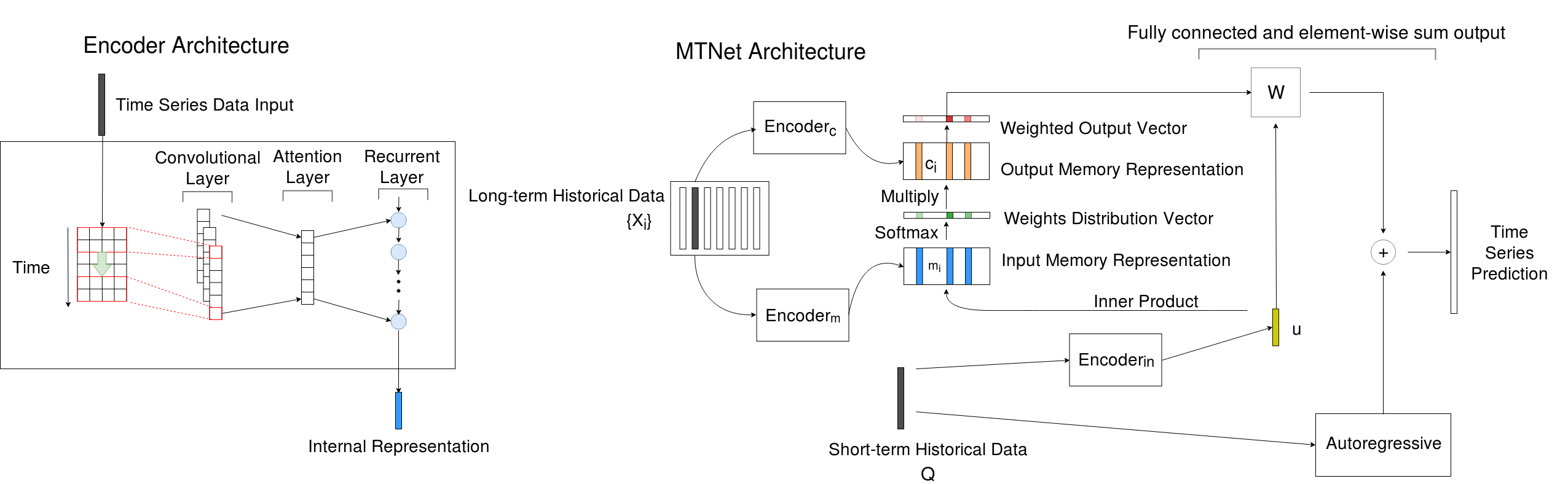}
}
\vspace{1mm}
\caption{An overview of Memory Time-series network (MTNet) on the right and the details of the encoder architecture on the left}
\label{fig:MTNet}
\end{figure*}



\section{Framework}
In this section, we first formulate time series forecasting problems and discuss the details of the proposed model MTNet.

\subsection{Problem Formulation}
We will formulate the task as multivariate time series forecasting since univariate forecasting is only a special case of it. Given a set of fully observed time series signals $\pmb{Y} = \{\pmb{y}_1, \pmb{y}_2, \ldots, \pmb{y}_T \}$ where $\pmb{y}_t \in \R^D$, and $D$ represents the variable dimension.
We aim at predicting series in a rolling forecasting fashion. That being said, to predict $\pmb{y}_{T+h}$, we assume $\{ \pmb{y}_1, \pmb{y}_2, \ldots, \pmb{y}_T\}$ are available. Likewise, to predict the  value of the next time stamp $\pmb{y}_{T+h+1}$, we assume $\{ \pmb{y}_1, \pmb{y}_2, \ldots, \pmb{y}_T, \pmb{y}_{T+1}\}$ are available. 

In most of the cases, the horizon of the forecasting task is chosen according to the demands of the environmental settings, e.g. for the traffic usage, the horizon of interest ranges from hours to a day; for the stock market data, even seconds/minutes-ahead forecast can be meaningful for generating returns.

\subsection{Memory Time-series Network}

Figure \ref{fig:MTNet} presents an overview of the proposed model MTNet architecture. MTNet takes a set of long-term time series historical data \(\{\pmb{X}_i\} = \pmb{X}_1,\cdots,\pmb{X}_n\) that are to be store in the memory and a short-term historical time series data $\pmb{Q}$ as input. Note that $\pmb{X}$ and $\pmb{Q}$ is not overlapped. 
The model writes ${\pmb{X}_i}$ to the fixed size memory, and then embeds $\pmb{X}$ and $\pmb{Q}$ into a fixed length internal representation using $\text{Encoder}_m$ and $\text{Encoder}_{in}$ respectively. Two separate encoders are used since our goal is not to find the most similar block to $\pmb{Q}$. We aim to attend to blocks stored in the memory that most likely resemble the ground truth to be predicted. This is further justified through our experiment. The attention weights, which symbolized the significance of different periods, is then derived by simply taking the inner product of their embedding. Then the model uses another $\text{Encoder}_c$ to obtain the context vector of the memory $\pmb{X}$ and multiply it with the attention weights to obtained the weighted memory vectors. MTNet then concatenates the weighted output vectors and the embedding of $\pmb{Q}$, and feed them as input to a dense layer to generate the outputs. In the end, the model generates the final prediction by summing up the outputs of the neural network with the outputs of the autoregressive model. The whole model can be trained end-to-end supervisedly. 

In the following sections, we introduce the details of building blocks in the MTNet.
\subsubsection{Encoder Architecture}
We use a convolutional layer without pooling to extract short-term patterns in the time dimension and local dependencies between variables. Let the input matrix be \(\pmb{X}\) where $\pmb{X} \in \R^{D \times T}$. The convolutional layer consists of multiple kernels with size \(w\) in time dimension and size \(D\) in variable dimension. The \(k\)-th filter sweeps through the input matrix \(\pmb{X}\), which can be formulated as,
\begin{align}
    \pmb{h}_k = \text{ReLU}(\pmb{W}_k \ast \pmb{X} + \pmb{b}_k)
\end{align}
where \(\ast\) denotes the convolutional operation, the output \(\pmb{h}_k\) would be a vector, and the ReLU function is \(\text{ReLU}(x) = max(0, x)\). The output of the convolutional layer is matrix with size \(d_c \times T_c\) where \(d_c\) denotes the number of filters and \(T_c = T - w + 1\) .

We apply an attention layer to convolutional layer's output matrix over the time dimension, That is, we can view the matrix as a sequence of \(d_c\)-dimensional vectors and the sequence length is \(T_c\). We apply attention over the time dimension so that our model can select relative time across all time steps adaptively.

The output of the attention layer is then fed into the recurrent component.  The recurrent component is a recurrent layer with the Gated Recurrent Unit (GRU) \cite{DBLP:journals/corr/ChungGCB14} and using the ReLU function as the hidden activation function. The hidden state of recurrent units at time \(t\) can be formulated as,
\begin{align}
    \pmb{r}_t &= \sigma(\pmb{x}_t\pmb{W}_{xr} + \pmb{h}_{t-1}\pmb{W}_{hr} + \pmb{b}_r) \\
    \pmb{u}_t &= \sigma(\pmb{x}_t\pmb{W}_{xu} + \pmb{h}_{t-1}\pmb{W}_{hu} + \pmb{b}_u) \\
    \pmb{c}_t &= \text{ReLU}(\pmb{x}_t\pmb{W}_{xc} + \pmb{r}_t\odot(\pmb{h}_{t-1}\pmb{W}_{hc}) + \pmb{b}_c) \\
    \pmb{h}_t &= (\mathbbm{1}-\pmb{u}_t)\odot \pmb{h}_{t-1} + \pmb{u}_t\odot \pmb{c}_t
\end{align}
where $\odot$ is the element-wise product, $\sigma$ is the sigmoid function and $\pmb{x}_t$ is the input of this layer at time $t$.
And the output is a fixed length internal representation.

\subsubsection{Input memory representation}
Suppose we are given a set of long-term historical data \(\{\pmb{X}_i\} = \pmb{X}_1,\cdots,\pmb{X}_n\) to be stored in memory where \(\pmb{X}_i \in \mathbb{R}^{D \times T}\). First, we fed \(\pmb{X}_i\) as input to the encoder and let the output be \(\pmb{m}_i\) where \(\pmb{m}_i \in \mathbb{R}^d\), which is the embedding of \(\pmb{X}_i\). Therefore, the entire set \(\{\pmb{X}_i\}\) are converted into input memory vectors \(\{\pmb{m}_i\}\) by embedding each \(\pmb{X}_i\).
And the short-term historical data \(\pmb{Q}\) is also embedded via another encoder to obtain its internal representation \(\pmb{u}\) where \(\pmb{Q} \in \mathbb{R}^{D \times T}\) and \(\pmb{u} \in \mathbb{R}^d\).
\begin{align}
    \pmb{m}_i &= \text{Encoder}_{m}(\it{\pmb{X}_i}) \\
    \pmb{u} &= \text{Encoder}_{in}(\it{\pmb{Q}})
\end{align}
where $\text{Encoder}_m$ and $\text{Encoder}_{in}$ are the encoders to convert $\pmb{X}_i$ and $\pmb{u}$. In the embedding space, we compute the match between \(\pmb{u}\) and each memory vector \(\pmb{m}_i\) by taking the inner product in the embedding space followed by a softmax. 
\begin{align}
    p_i = \text{Softmax}(\pmb{u}^\top\pmb{m}_i)
\end{align}
where the softmax is \(\text{Softmax}(z) = e^{z_i} / \sum_j e^{z_j}\) and we can define a vector $\pmb{p}$ be viewed as the attention weights distribution vector over the memory inputs.

\subsubsection{Output memory representation}
Each $\pmb{X}_i$ in the long-term historic time series data has a corresponding output vector $\pmb{c}_i$ obtained via another encoder $\text{Encoder}_{c}$. $\pmb{c}_i$ corresponds to the context vector as in Natural Language Processing.
\begin{align}
    \pmb{c}_i = \text{Encoder}_c(\pmb{X}_i)
\end{align}
where \(\pmb{c}_i \in \mathbb{R}^d\). And each \(\pmb{c}_i\) in the output set has a corresponding weighted output vector $o_i$.
\begin{align}
    \pmb{o}_i = p_i \times \pmb{c}_i
\end{align}
where \(\pmb{o}_i \in \mathbb{R}^d\).

\subsubsection{Autoregressive Component}
The non-linearity of both Convolutional and Recurrent layer causes the scale of neural network outputs to be insensitive. To address this drawback, we decompose the final prediction of MTNet into a linear part and a non-linear part which is in similar inspirit to the highway network \cite{srivastava2015highway} and LSTNet.
We use the classical Autoregressive(AR) model as the linear component as in \cite{lai2017modeling}. Denote the forecasting result of the AR component at time stamp \(t\) as \(\pmb{Y}_t^L \in \mathbb{R}^D\), \(s^{ar}\) as the size of the input window, the coefficient of the AR model as \(\pmb{w}^{ar} \in \mathbb{R}^{s^{ar}}\) and \(b^{ar} \in \mathbb{R}\). In our model, all dimension variables share the same set of linear parameters. The AR model is formulated as follows,
\begin{align}
    \pmb{y}^L_{t,i} = \sum_{k = 0}^{s^{ar} - 1}\pmb{w}_{k}^{ar}\pmb{q}_{t - k,i} + b^{ar}
\end{align}

\subsubsection{Generating the prediction}
We use a dense layer to combine the internal representation of short-term historical data \(\pmb{u}\) and weighted output vector set \(\{\pmb{o}_i\}\). The output of the dense layer is computed as, 
\begin{align}
    \pmb{y}_t^D = \pmb{W}^D[\pmb{u};\pmb{o}_1;\pmb{o}_2;\cdots;\pmb{o}_T]+ \pmb{b}
\end{align}
where $[\pmb{u};\pmb{o}_1;\pmb{o}_2;\cdots;\pmb{o}_T]$ is the concatenation of the internal representation $\pmb{u}$ and the output vector set \(\{\pmb{o}_i\}\), and \(\pmb{y}^D_y \in \mathbb{R}^D\) is the prediction of the neural network part.

The final prediction of MTNet is obtain by integrating the outputs of the neural network part and the AR component,
\begin{align}
    \hat{\pmb{y}}_t = \pmb{y}^D_t + \pmb{y}^L_t
\end{align}
 where $\hat{\pmb{y}}_t$ denotes the model's final prediction at time stamp \(t\).

\subsubsection{Objective function}
In the training process, we adopt the mean absolute error (L1-loss) and the objective function:
\begin{align}
    \mathcal{O}(\pmb{y}_T, \hat{\pmb{y}}_T) = \frac{1}{N}\sum_{j=1}^N\sum_{i=1}^D\lvert(\hat{{\pmb{y}}}^j_{T,i} - \pmb{y}_{T,i}^j)\rvert
\end{align}
where \(N\) is the number of training samples and \(D\) is the dimension of target data. All neural models are trained using the Adam optimizer \cite{DBLP:journals/corr/KingmaB14}.
\section{Experiments}

We conducted extensive experiments with 8 methods (including our
new methods) on 6 benchmark datasets for both univariate and multivariate time series forecasting tasks. All the data are available online. 

\begin{table}
	\begin{center}
        \resizebox{\columnwidth}{!}{%
		\begin{tabular}{lrrrrrrrrr} 
		\toprule
		Datasets 		& $T$		& $D$	&$K$ & $L$		 \\
		\midrule
		Beijing PM2.5 & 43824 & 7 & 1 & 1 hour 
		\\
		GefCom Electricity Price & 21552 & 9 & 1 & 1 hour 
	    \\
		Traffic 		& 17,544 	& 862 	&862& 1 hour	\\
        Solar-Energy			& 52,560	& 137&137	& 10 minutes\\
		Electricity 	& 26,304 	& 321 &321	& 1 hour	\\
        Exchange-Rate		& 7,588		& 8	&8	& 1 day		\\
        \bottomrule
		\end{tabular}
        }
       
	\end{center}
	\caption{Dataset Statistics, where $T$ is length of time series, $D$ is the total number of variables, $K$ is the number of variables to be predicted, $L$ is the sample rate.}
	    \label{data-stats}

\end{table}

\subsection{Datasets}
We chose 6 publicly available benchmark datasets, including 2 univariate datasets: Beijing PM2.5\footnote[1]{\href{https://archive.ics.uci.edu/ml/datasets.html}{https://archive.ics.uci.edu/ml/datasets.html}}, GEFCom(2014) Electricity Price \cite{hong2016probabilistic}, and 4 multivariate datasets \cite{lai2017modeling}: Traffic, Solar-Energy, Electricity, Exchange-Rate.
Table \ref{data-stats} summarizes the corpus statistics.

\begin{table*}[h]

\centering
\resizebox{.6\textwidth}{!}{
\centering
\begin{tabular}{cc|cccc|cccc}
 \\[-1em]
\\[-1em]                       
\hline
\\[-1em]
\\[-1em]
\multicolumn{2}{c|}{\multirow{2}{*}{Dataset}} & \multicolumn{4}{c|}{\multirow{2}{*}{Beijing PM2.5 Dataset}} & \multicolumn{4}{c}{\multirow{2}{*}{GefCom Electricity(2014)}} \\
\multicolumn{2}{c|}{}                         & \multicolumn{4}{c|}{}                                       & \multicolumn{4}{c}{}    \\               \\[-1em]
\\[-1em]                       
\hline
\\[-1em]
\\[-1em]
\multicolumn{2}{c|}{}                         & \multicolumn{4}{c|}{horizon}                                & \multicolumn{4}{c}{horizon}            \\[-1em]
\\[-1em]                       \\ 
\hline
\\[-1em]
\\[-1em]
Methods                      & Metrics         & 3             & 6             & 12           & 24           & 3              & 6             & 12            & 24            \\ 
\\[-1em]
\\[-1em]
\hline
\\[-1em]
\\[-1em]
\multirow{2}{*}{AR}         
& MAE    & 24.94   & 37.49  & 50.60  & 61.53        
         & 5.84    & 7.70  & 8.82  &    9.40      \\
& RMSE   & \textbf{38.40}${^\dagger}$    & 54.03  & \textbf{69.77}${^\dagger}$  & \textbf{82.57}${^\dagger}$        
         &  10.34   & 13.16  & 15.51  & 16.80         \\
\\[-1em]
\\[-1em] 
\hline
\\[-1em]
\\[-1em]
         \multirow{2}{*}{LRidge}         
& MAE    & 24.66    & 36.65  & 49.55  & 61.03        
         & 5.52    & 7.32  & 8.55  & 9.39         \\
& RMSE   & 39.23    & 54.81  & 70.53  & 82.58        
         & 9.78    & 12.58  & 14.74  & 16.59         \\
\\[-1em]
\\[-1em] 
\hline
\\[-1em]
\\[-1em]
         \multirow{2}{*}{LSVR}         
& MAE    & 24.28    &  35.93 & 48.65  & 60.32        
         & 7.52    & 8.67  & 9.40  &    9.40      \\
& RMSE   & 39.28    & 55.13  & 70.86  & 83.38        
         & 13.20    & 14.91  & 15.74  & \textbf{16.43}$^{\dagger}$         \\
         
    \\[-1em]
\\[-1em] 
\hline
\\[-1em]
\\[-1em]
         \multirow{2}{*}{GP}         
& MAE    & 23.89    & 36.98  & 51.84  & 64.75        
         & 12.33    & 15.33  & 14.75  & 12.13         \\
& RMSE   & 40.06    & 58.58  & 77.24  & 91.03        
         & 22.91    & 26.99  & 26.66  & 23.05         \\
         \\[-1em]
\\[-1em]
         \hline
        \\[-1em]
\\[-1em]
         \multirow{2}{*}{VARMLP}         
& MAE    & 90.70    & 88.70  & 82.57  & 74.58        
         & 26.76    & 24.23  & 24.93  & 23.71         \\
& RMSE   & 126.3    & 122.6  & 113.6  & 103.1        
         & 42.39    & 39.34  & 40.02  & 39.43         \\
        \\[-1em]
\\[-1em] 
         \hline
         \\[-1em]
\\[-1em]
         \multirow{2}{*}{RNN-GRU}         
& MAE    & 81.76    & 78.71  & 73.72  & 68.07        
         & 23.61    & 22.86  & 23.21  & 23.50         \\
& RMSE   & 115.0    & 110.2  & 102.30 & 96.86        
         & 39.03    & 39.28  & 39.33  & 39.42           \\
         \\[-1em]
\\[-1em]
         \hline
         \\[-1em]
\\[-1em]
         \multirow{2}{*}{DA-RNN}         
& MAE    & 26.51    & 39.59  & 52.54  & 60.89        
         & 8.41    & 10.24  & 10.61  & 9.79        \\
& RMSE   & 42.07    & 58.82  & 73.31  & 84.30        
         & 15.02    & 18.31 & 19.35  & 17.84         \\
        \\[-1em]
\\[-1em] 
         \hline
         \hline
         \\[-1em]
\\[-1em]
         \multirow{2}{*}{MTNet}         
& MAE    & \textbf{23.16}$^{\dagger}$    & \textbf{34.10}$^{\dagger}$  & \textbf{48.75}$^{\dagger}$  & \textbf{60.55}$^{\dagger}$       
         & \textbf{4.99}$^{\dagger}$    & \textbf{6.25}$^{\dagger}$  & \textbf{7.76}$^{\dagger}$  & \textbf{9.12}$^{\dagger}$         \\
         
& RMSE   & 38.52    & \textbf{53.69}$^{\dagger}$  & 73.72  & 85.93 
         & \textbf{9.47}$^{\dagger}$    & \textbf{11.12}$^{\dagger}$  & \textbf{14.11}$^{\dagger}$ & 16.61        \\
        \\[-1em]
\\[-1em] 
         \hline
\\[-1em]
\\[-1em]
\end{tabular}

}
\caption{Time series prediction results over different univariate datasets (best performance is displayed in \textbf{boldface} in each case). Best performance is labelled with $\dagger$ if P-value of the t-test is lower than 0.05.}

\label{table:2}
\end{table*}

\subsubsection{univariate}
\begin{itemize}
    \item Beijing PM2.5: It contains hourly PM2.5 data and the associated meteorological data in Beijing of China. PM2.5 measurement is the target series. The exogenous time series include dew point, temperature, pressure, combined wind direction, cumulated wind speed, hours of snow, and hours of rain. Totally we have 43,824 multivariable sequences. 
    \item GEFCom2014 Electricity Price: The datasets were published in GEFCom2014 forecasting competition \cite{hong2016probabilistic}. The competition contained four problems: electricity load forecasting, electricity price forecasting, and two other problems related to wind and solar power generation. We chose the electricity price forecasting problem since it is the only univariate task out of the four. Additional to all features given, we also added calender-based features such as hour of the day, day of the week, and day of the year. The number of sequence is 21552. 
\end{itemize}
\subsubsection{multivariate}
\begin{itemize}
    \item Traffic: A collection of 48 month (2015-2016) hourly data from California Department of Transportation. The data describes the road occupancy rates (between 0 and 1) measured by different sensor in San Francisco Bay area freeways.
    \item Solar-Energy: The solar power production records in the year 2006, which is sampled every 10 minutes from 137 PV plants in Alabama State.
    \item Electricity: The electricity consumption in kWh was recorded every 15 minutes from the year 2012 to 2014 for 321 clients. We convert the data to represent hourly consumption.
    \item Exchange-Rate: the collection of the daily exchange rates of eight foreign countries including Australia, British, Canada, Switzerland, China, Japan, New Zealand and Singapore ranging from 1990 to 2016.
\end{itemize}
In our experiments, all datasets have been split into training set (60\%), validation set
(20\%) and test set (20\%) in chronological order.

\subsection{Methods for comparison}
The methods in our comparative evaluations are the follows. 
\begin{itemize}
    \item AR: The autogressive model, which is equivalent to the one-dimensional vector autoregressive (VAR) model.
    \item LRidge: The autoregressive model where the loss function is the linear least squares function and regularization is given by the l2-norm.
    \item LSVR: The autoregressive model with Support Vector Regression objective function \cite{NIPS1996_1187}.
    \item GP: The Gaussian Process time series model \cite{roberts2013gaussian,frigola2015bayesian}.
    \item VAR-MLP: The model combines the Multilayer Perception (MLP) and autoregressive model (VAR) \cite{zhang2003time}.
    \item RNN-GRU: The recurrent neural network model using GRU cell for time series forecasting.
    \item DA-RNN: The dual-stage attention based recurrent neural network \cite{qin2017dual}. The attention mechanism in the first stage learns the weights of input variables, and the attention mechanism in the second stage learns the weights of hidden states across all time steps for forecasting.
    \item LSTNet: Long- and Short-term Time-series Network \cite{guo2018multi}. This model contains convolutional layer to extract the local dependency patterns, recurrent layer to capture long-term dependency patterns and recurrent-skip layer to capture periodic properties in the input data for forecasting .
\end{itemize}

\subsection{Metrics}
We consider root mean squared error (RMSE) and mean absolute error (MAE) for univariate task and root relative squared error (RRSE) and empirical correlation coefficient (CORR) for multivariate task. All the metrics are commonly used in the corresponding tasks. These four metrics are defined as:
\begin{itemize}
    \item Root Mean Squared Error (RMSE):
    \begin{align}
        \text{RMSE} = \sqrt{\frac{1}{N}\sum_{i = 1}^N(\pmb{y}_t^i - \hat{\pmb{y}}_t^i)^2}
    \end{align}
    \item Mean Absolute Error (MAE):
    \begin{align}
        \text{MAE} = \frac{1}{N}\sum_{i = 1}^N\lvert\pmb{y}_t^i - \hat{\pmb{y}}_t^i\rvert
    \end{align}
    \item Root Relative Squared Error (RRSE):
    \begin{align}
        \text{RRSE} = \frac{\sqrt{\sum_{(i,t) \in \Omega_{\text{Test}}(\pmb{Y}_{i,t} - \hat{\pmb{Y}}_{i,t})^2}}}{\sqrt{\sum_{(i,t) \in \Omega_{\text{Test}}(\pmb{Y}_{i,t} - \text{mean}(\pmb{Y}))^2}}}
    \end{align}
    \item Empirical Correlation Coefficient (CORR): 
    \begin{multline}
     \text{CORR} = \\ \frac{1}{n}\sum_{i = 1}^n\frac{\sum_t(\pmb{Y}_{i,t} - \text{mean}(\pmb{Y}_i))(\hat{\pmb{Y}}_{i,t} - \text{mean}(\hat{\pmb{Y}}_i))}{\sqrt{\sum_t(\pmb{Y}_{i,t} - \text{mean}(\pmb{Y}_i))^2(\hat{\pmb{Y}}_{i,t} - \text{mean}(\hat{\pmb{Y}}_i))^2}}
    \end{multline}
\end{itemize}
where \(\pmb{y}_t^i,\hat{\pmb{y}}_t^i \in \mathbb{R}\) are ground true values and model predictions in univariate task. \(\textbf{Y}, \hat{\textbf{Y}} \in \mathbb{R}^{n \times T}\) are ground true values and model predictions in multivariate task. RSE is the scaled version of Root Mean Squared Error (RMSE), which is designed to make the evaluation more readable regardless the data scale, and $\emph{$\Omega$}_{Test}$ is the set of time stamps used for testing. For MAE, RMSE and RSE lower value is better, while higher value is better for CORR.

\begin{table*}[h]
\centering
\resizebox{\textwidth}{!}{

\begin{tabular}{ll|cccc|cccc|cccc|cccc}

\hline
\\[-1em]
\\[-1em]
\\[-1em]
\multicolumn{2}{c|}{Dataset} & \multicolumn{4}{|c|}{Solar-Energy} & \multicolumn{4}{|c|}{Traffic} & \multicolumn{4}{|c|}{Electricity} & \multicolumn{4}{|c}{Exchange-Rate}
\\
\\[-1em]
\\[-1em]
\\[-1em]
\hline
\\[-1em]
\\[-1em]
\\[-1em]
\multicolumn{2}{c|}{} & \multicolumn{4}{|c|}{Horizon} & \multicolumn{4}{|c|}{Horizon} & \multicolumn{4}{|c|}{Horizon} & \multicolumn{4}{|c}{Horizon}  \\
\\[-1em]
\\[-1em]
\\[-1em]
\hline
\\[-1em]
\\[-1em]
\\[-1em]

Methods & Metrics & 3 & 6 & 12 & 24 & 3 & 6 & 12 & 24 & 3 & 6 & 12 & 24 & 3 & 6 & 12 & 24 \\
\\[-1em]
\\[-1em]
\\[-1em]
\hline
\\[-1em]
\\[-1em]
\\[-1em]
AR&RSE&0.2435&0.3790&0.5911&0.8699&0.5991&0.6218&0.6252&0.6293&0.0995&0.1035&0.1050&0.1054&0.0228&0.0279&0.0353&0.0445\\

&CORR&0.9710&0.9263&0.8107&0.5314&0.7752&0.7568&0.7544&0.7519&0.8845&0.8632&0.8591&0.8595&0.9734&0.9656&0.9526&0.9357\\
\\[-1em]
\\[-1em]
\\[-1em]
\hline
\\[-1em]
\\[-1em]
\\[-1em]
LRidge&RSE&0.2019&0.2954&0.4832&0.7287&0.5833&0.5920&0.6148&0.6025&0.1467&0.1419&0.2129&0.1280&\textbf{0.0184}$^{\dagger}$&0.0274&0.0419&0.0675\\

&CORR&0.9807&0.9568&0.8765&0.6803&0.8038&0.8051&0.7879&0.7862&0.8890&0.8594&0.8003&0.8806&\textbf{0.9788}$^{\dagger}$&\textbf{0.9722}$^{\dagger}$&0.9543&0.9305\\
\\[-1em]
\\[-1em]
\\[-1em]
\hline
\\[-1em]
\\[-1em]
\\[-1em]
LSVR&RSE&0.2021&0.2999&0.4846&0.7300&0.5740&0.6580&0.7714&0.5909&0.1523&0.1372&0.1333&0.1180&0.0189&0.0284&0.0425&0.0662\\

&CORR&0.9807&0.9562&0.8764&0.6789&0.7993&0.7267&0.6671&0.7850&0.8888&0.8861&0.8961&0.8891&0.9782&0.9697&0.9546&0.9370\\
\\[-1em]
\\[-1em]
\\[-1em]
\hline
\\[-1em]
\\[-1em]
\\[-1em]

GP&RSE&0.2259&0.3286&0.5200&0.7973&0.6082&0.6772&0.6406&0.5995&0.1500&0.1907&0.1621&0.1273&0.0239&0.0272&0.0394&0.0580\\

&CORR&0.9751&0.9448&0.8518&0.5971&0.7831&0.7406&0.7671&0.7909&0.8670&0.8334&0.8394&0.8818&0.8713&0.8193&0.8484&0.8278\\
\\[-1em]
\\[-1em]
\\[-1em]
\hline
\\[-1em]
\\[-1em]
\\[-1em]

VARMLP&RSE&0.1922&0.2679&0.4244&0.6841&0.5582&0.6579&0.6023&0.6146&0.1393&0.1620&0.1557&0.1274&0.0265&0.0394&0.0407&0.0578\\
&CORR&0.9829&0.9655&0.9058&0.7149&0.8245&0.7695&0.7929&0.7891&0.8708&0.8389&0.8192&0.8679&0.8609&0.8725&0.8280&0.7675\\

\\[-1em]
\\[-1em]
\\[-1em]
\hline
\\[-1em]
\\[-1em]
\\[-1em]

RNN-GRU&RSE&0.1932&0.2628&0.4163&0.4852&0.5358&0.5522&0.5562&0.5633&0.1102&0.1144&0.1183&0.1295&0.0192&0.0264&0.0408&0.0626\\
&CORR&0.9823&0.9675&0.9150&0.8823&0.8511&0.8405&0.8345&0.8300&0.8597&0.8623&0.8472&0.8651&0.09786&0.9712&0.9531&0.9223\\
\\[-1em]
\\[-1em]
\\[-1em]
\hline
\\[-1em]
\\[-1em]
\\[-1em]
LSTNet	& RSE 
							  & 0.1916 & 0.2475 & 0.3449 & 0.4521
							  & 0.4818 & 0.4920 & 0.4948 & \textbf{0.5048}$^{\dagger}$
                			  & 0.0900 & 0.0997 & 0.1040 & 0.1444
							  & 0.0226 & 0.0280 & 0.0356 & 0.0449\\
           & CORR 
                       		  & 0.9820 & 0.9698 & 0.9394 & 0.8911 & 0.8725 & 0.8672 & 0.8582 & \textbf{0.8617}$^{\dagger}$
                              & 0.9305 & 0.9153 & 0.9137 & 0.9125
                              & 0.9735 & 0.9658 & 0.9511 & 0.9354\\ 
\\[-1em]
\\[-1em]
\\[-1em]
\hline
\hline
\\[-1em]
\\[-1em]
\\[-1em]

MTNet&RSE&\textbf{0.1847}$^{\dagger}$&\textbf{0.2398}$^{\dagger}$&\textbf{0.3251}$^{\dagger}$&\textbf{0.4285}$^{\dagger}$&\textbf{0.4764}$^{\dagger}$&\textbf{0.4855}$^{\dagger}$&\textbf{0.4877}$^{\dagger}$&0.5023&\textbf{0.0840}$^{\dagger}$&\textbf{0.0901}$^{\dagger}$&\textbf{0.0934}$^{\dagger}$&\textbf{0.0969}$^{\dagger}$&0.0212&\textbf{0.0258}$^{\dagger}$&\textbf{0.0347}$^{\dagger}$&\textbf{0.0442}$^{\dagger}$\\
&CORR&\textbf{0.9840}$^{\dagger}$&\textbf{0.9723}$^{\dagger}$&\textbf{0.9462}$^{\dagger}$&\textbf{0.9013}$^{\dagger}$&\textbf{0.8728}&\textbf{0.8681}&\textbf{0.8644}$^{\dagger}$&0.8570&\textbf{0.9319}$^{\dagger}$&\textbf{0.9226}$^{\dagger}$&\textbf{0.9165}$^{\dagger}$&\textbf{0.9147}$^{\dagger}$&0.9767&0.9703&\textbf{0.9561}$^{\dagger}$&\textbf{0.9388}$^{\dagger}$\\
\end{tabular}
}
\caption{Time series prediction results over different multivariate datasets (best performance is displayed in \textbf{boldface} in each case). Best performance is labelled with $\dagger$ if P-value of the t-test is lower than 0.05.}
\label{table:3}

\end{table*}

\subsection{Experimental Details}
We conducted grid search over all tunable hyper-parameters on the validation set in each dataset for all methods. To be more specifically, all models chose input length from \(\{2^0,2^1,...,2^9\}\). For LRidge and LSVR, the regularization coefficient \(\lambda\) is chosen from \(\{2^{-10},2^{-8},...,2^8,2^{10}\}\). For GP, the RBF kernel bandwidth \(\sigma\) and the noise level \(\alpha\) are in the range of \(\{2^{-10},2^{-8},...,2^8,2^{10}\}\). For VARMLP, we conduct grid search over \(\{32, 50, 100\}\) for the size of dense layer. For RNN-GRU and DA-RNN, we conduct grid search over \(\{32, 50, 100\}\) for the hidden state size. For LSTNet, we conduct grid search over \(\{32, 50, 100\}\) for the hidden dimension size of recurrent layers and convolutional layer, \(\{20, 50, 100\}\) for recurrent-skip layers. For MTNet, we conduct grid search over \(\{32, 50, 100\}\) for the hidden dimension size of GRU and convolutional layer and we set the number of the long-term historical data series $n$ as 7. We use dropout after each layer except the input of each long-term historical data to input to the convolutional layer, and output in MTNet. The dropout rate is set to 0.2.

\subsection{Result: Time Series Prediction}
Table \ref{table:2} summarizes the results of univariate testing sets in the metrics MAE and RMSE. Table \ref{table:3} summarizes the results on multivariate testing sets in the metrics RRSE and CORR. We set \(horizon = \{3,6,12,24\}\), which means that the horizons were set from 3 to 24 hours for forecasting over the Beijing PM2.5, GEFCom2014 Price, Electricity and Traffic data, from 30 to 240 minutes over the Solar-Energy dataset, and from 3 to 24 days  over the Exchange-Rate dataset. 
Larger horizons make the forecasting harder. The best result for each data and metric pair is highlighted in boldface and with labelled with $\dagger$ if the p-value of t-test (comparing with the 2nd best) is lower than 0.05. Both tables show that our model outperforms other competitors significantly. 

\begin{figure}[!th]
\centering
\includegraphics[width=0.45\textwidth]{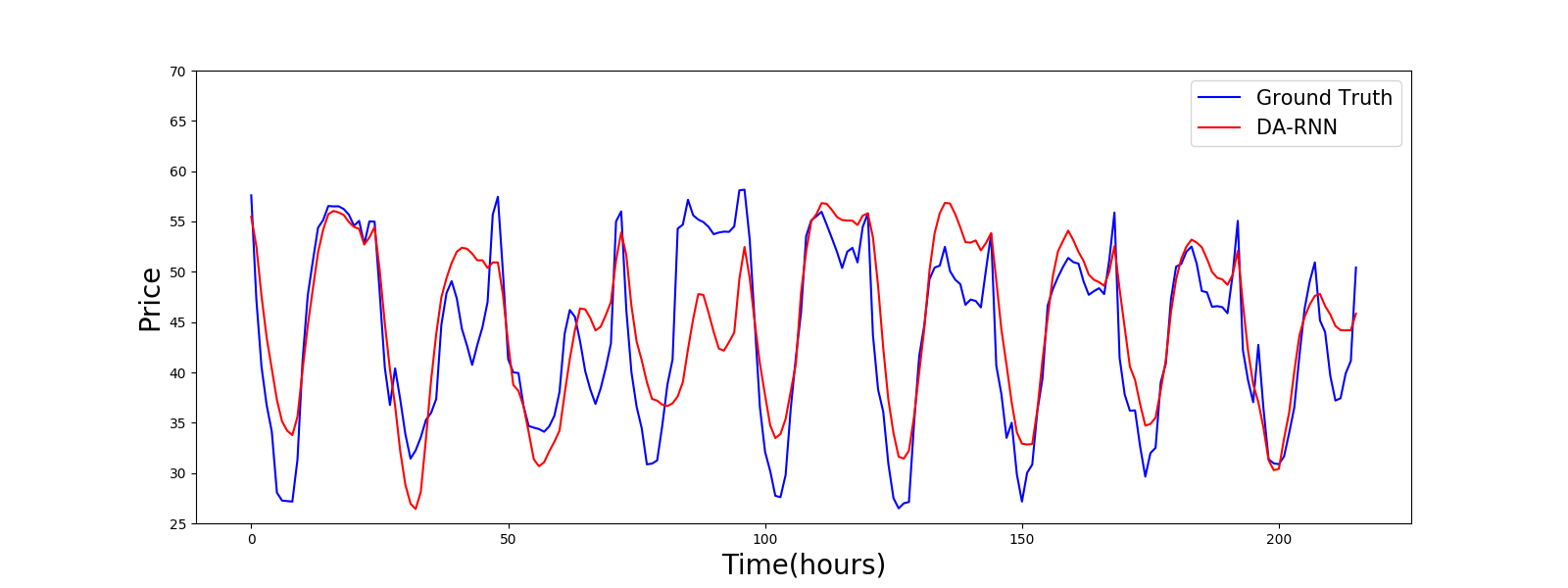}
\includegraphics[width=0.45\textwidth]{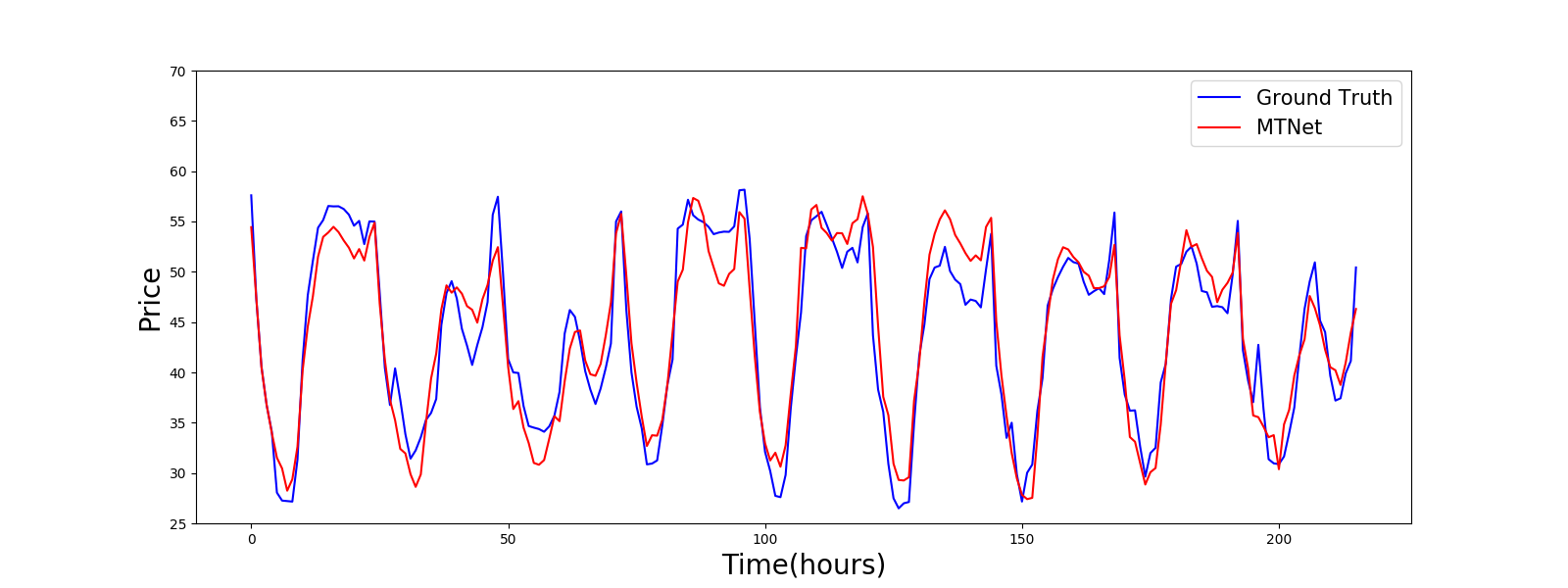}
\includegraphics[width=0.45\textwidth]{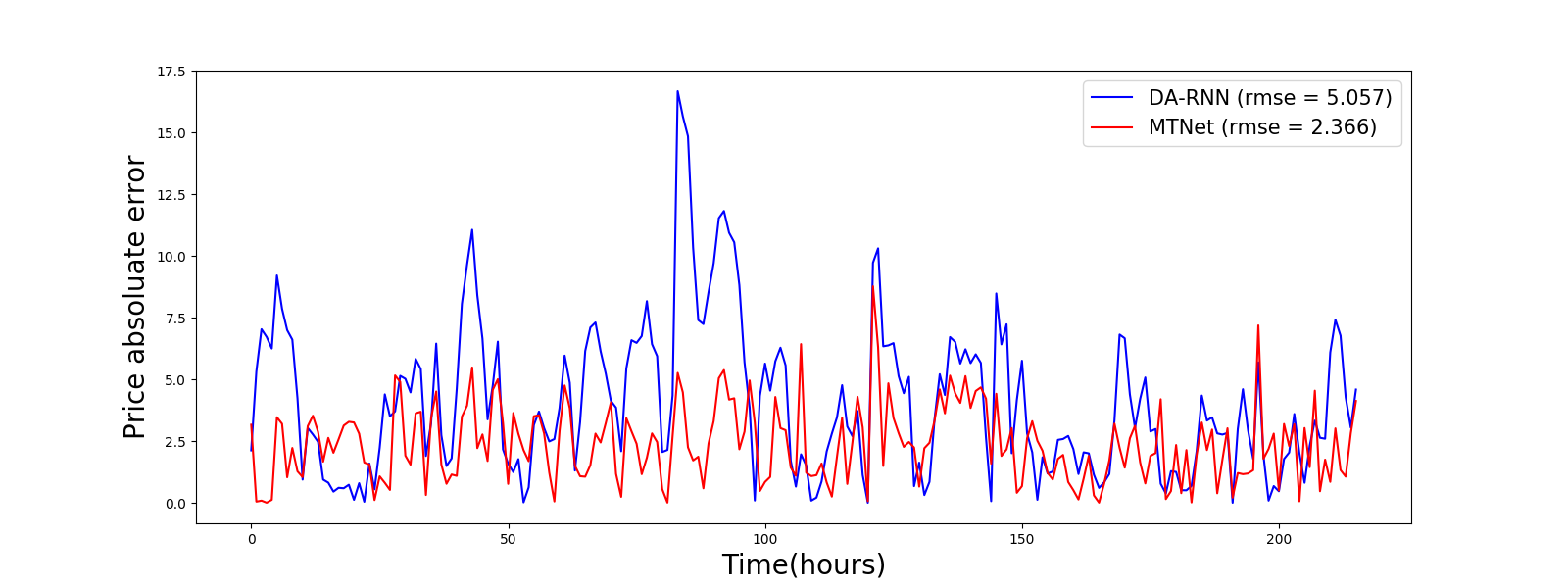}
\caption{Prediction results of DA-RNN and MTNet GEFCom2014 Electricity Price dataset visualized. Segmnets are randomly sampled from the testing set.}
\label{gef 24}
\end{figure}

\begin{figure}[!h]
\centering
\includegraphics[width=0.45\textwidth]{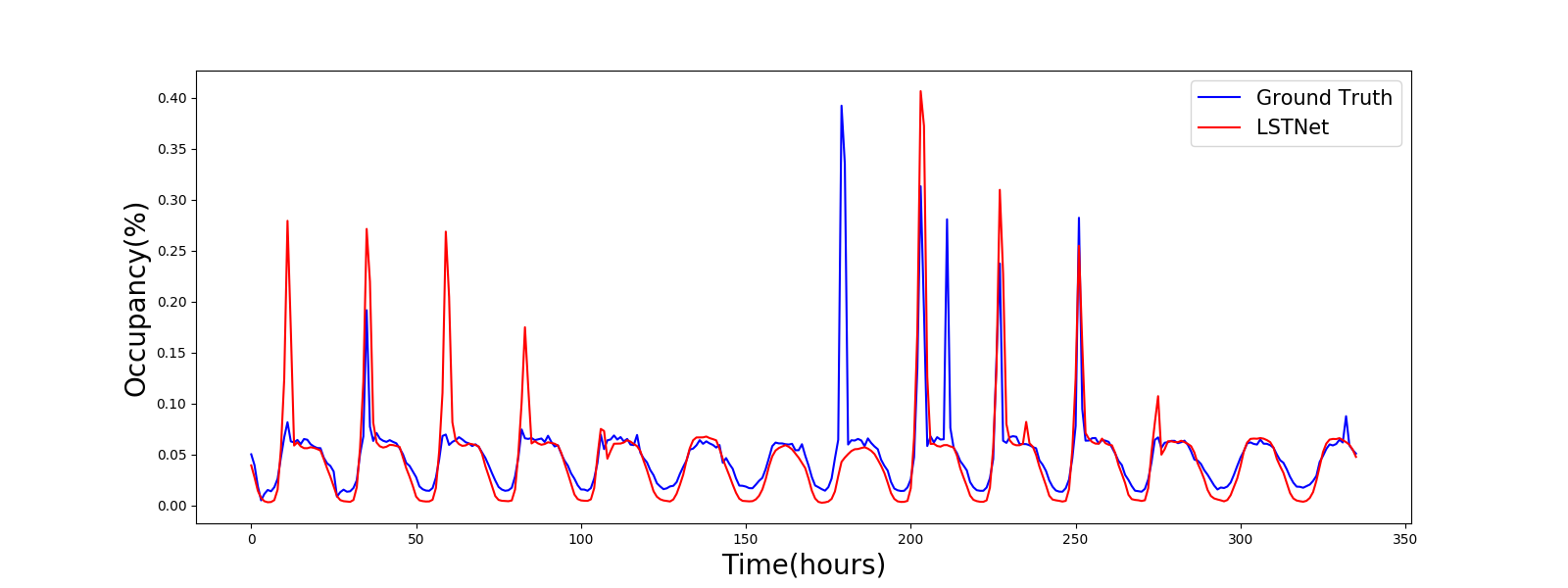}
\includegraphics[width=0.45\textwidth]{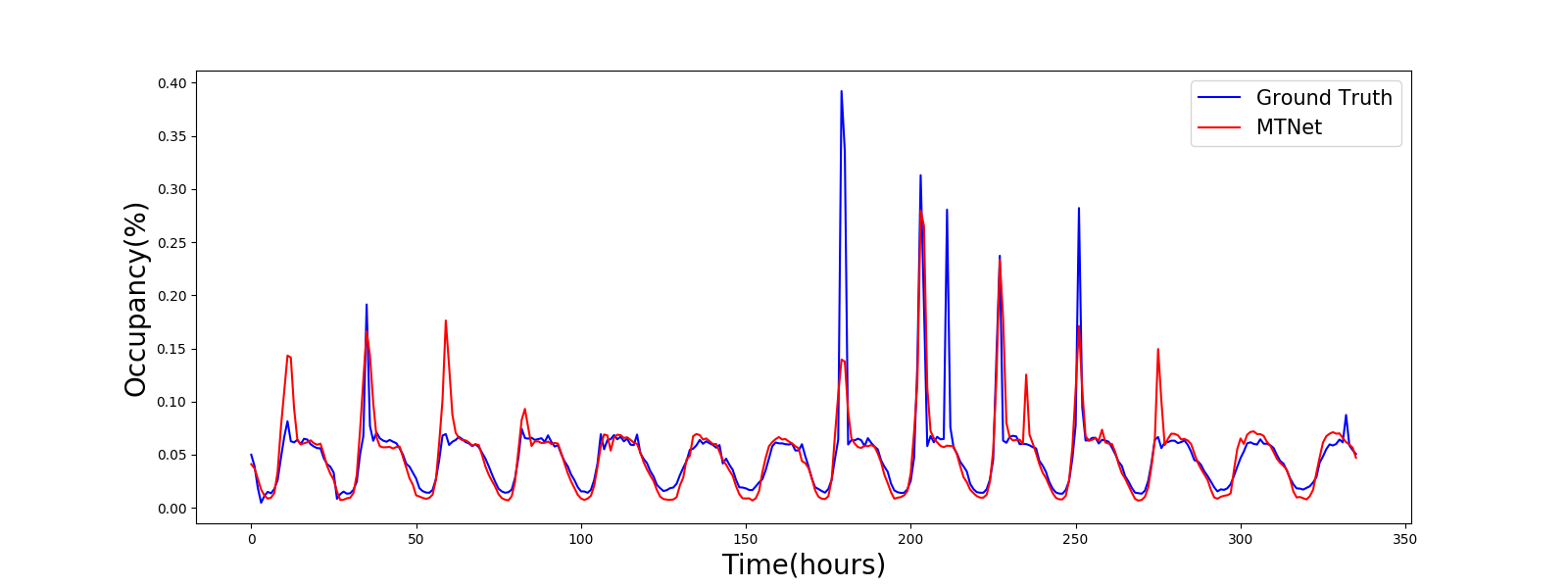}
\includegraphics[width=0.45\textwidth]{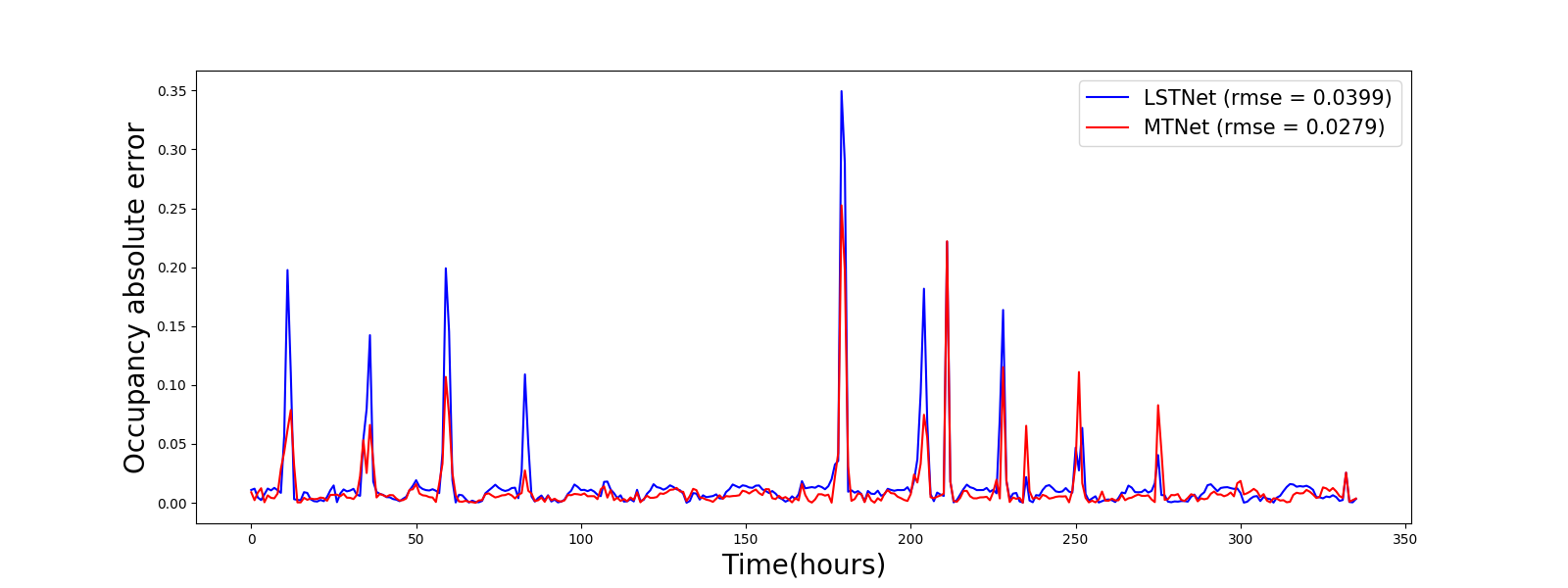}

\caption{Prediction results of LSTNet and MTNet on Traffic dataset with horizon 24 visualized. Segments are randomly sampled from the testing set}
\label{tra:24}
\end{figure}

For visualization, we also compare the prediction result of DA-RNN (2nd best solution for univariate forecasting) and MTNet over the GEFCom2014 Price dataset in Figure \ref{gef 24} and the prediction result of LSTNet and MTNet over the Traffic occupation dataset in Figure \ref{tra:24}. Blue lines are the ground truth time series and red lines are the prediction in the top two subfigures. The bottom subfigure illustrate the absolute error of the two compared models across every timestamps. 
We can observe that MTNet generally fits the ground truth much better than DA-RNN since the absolute error of MTNet is generally lower. The results show that our model produce more accurate predictions especially around peak values.

DA-RNN uses an input attention mechanism in the first stage to extract relevant input features, however, it can not capture the correlations between different components. In MTNet, we utilize convolutional components to learn the interactions between different variable dimensions. 
Furthermore, instead of utilizing a temporal attention layer to select relevant timesteps as in DA-RNN and LSTNet, we employed a different strategy. MTNet first transforms latest short-term data into fixed length representations and calculate attention weights with chunks of long-term historical data, which is also transformed into its embedding but through a different encoder. Intuitively, MTNet learns what \emph{period of time} is supportive to the prediction and thus is able to generate better prediction on datasets with periodical patterns.

\subsection{Interpretability of MTNet}
\begin{figure}[!bh]
\centering
   
   \includegraphics[width=0.45\textwidth]{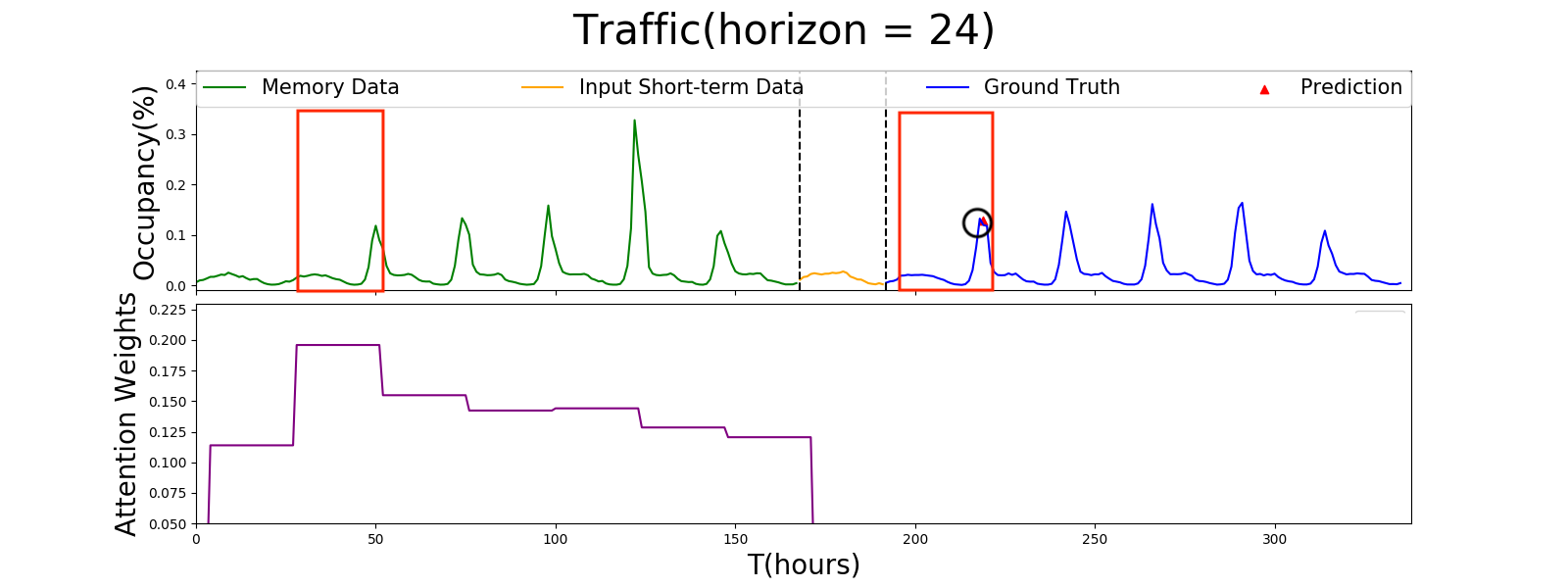}
   \vspace{1mm}
   
   \includegraphics[width=0.45\textwidth]{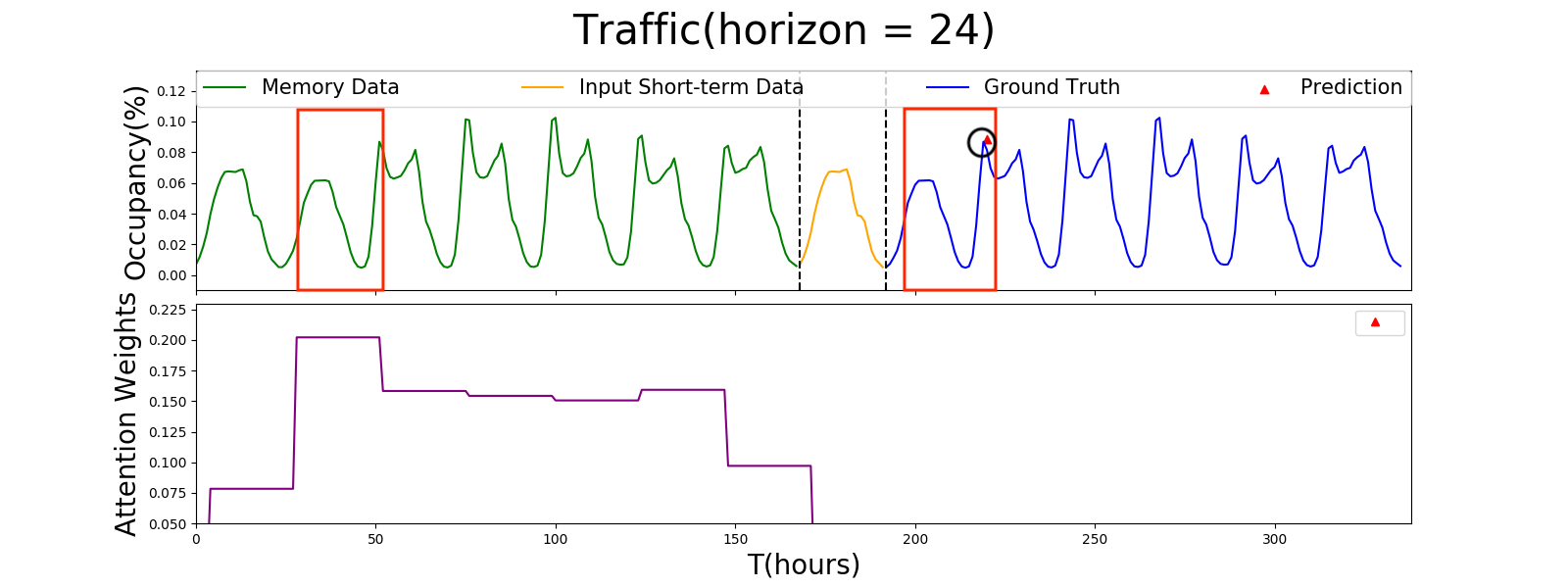}
      \vspace{1mm}

   \includegraphics[width=0.45\textwidth]{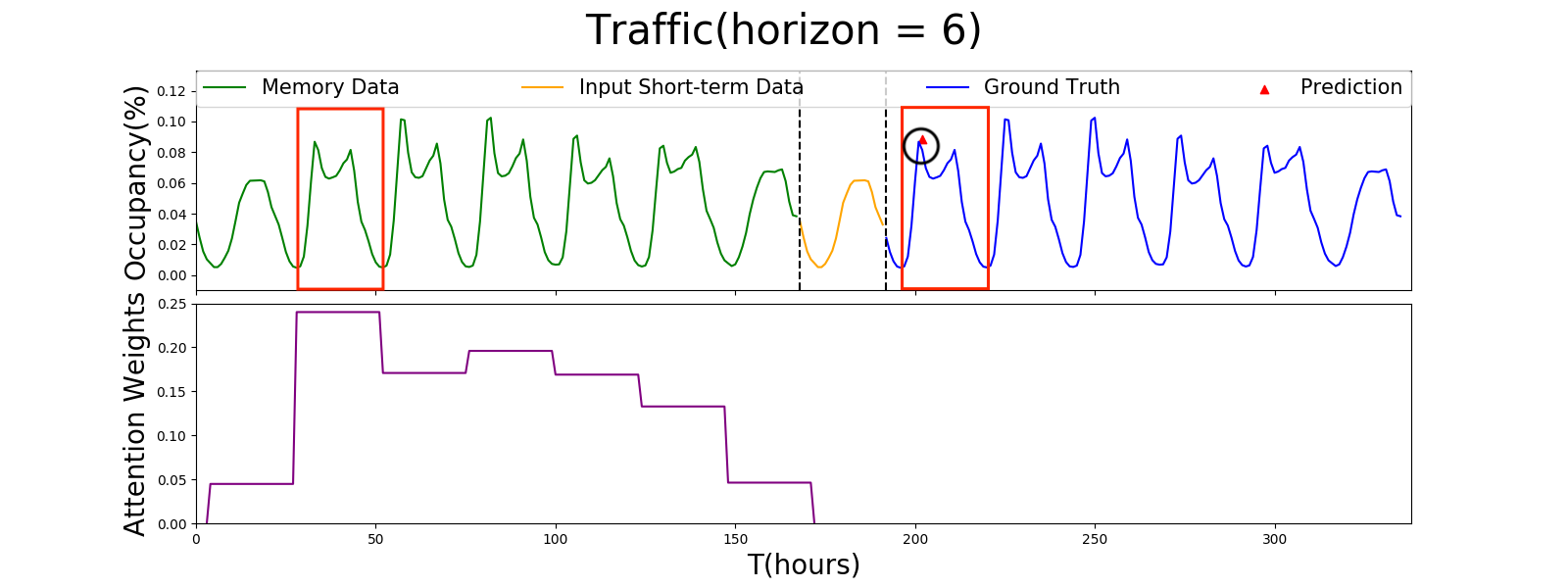}
      \vspace{1mm}

   \includegraphics[width=0.45\textwidth]{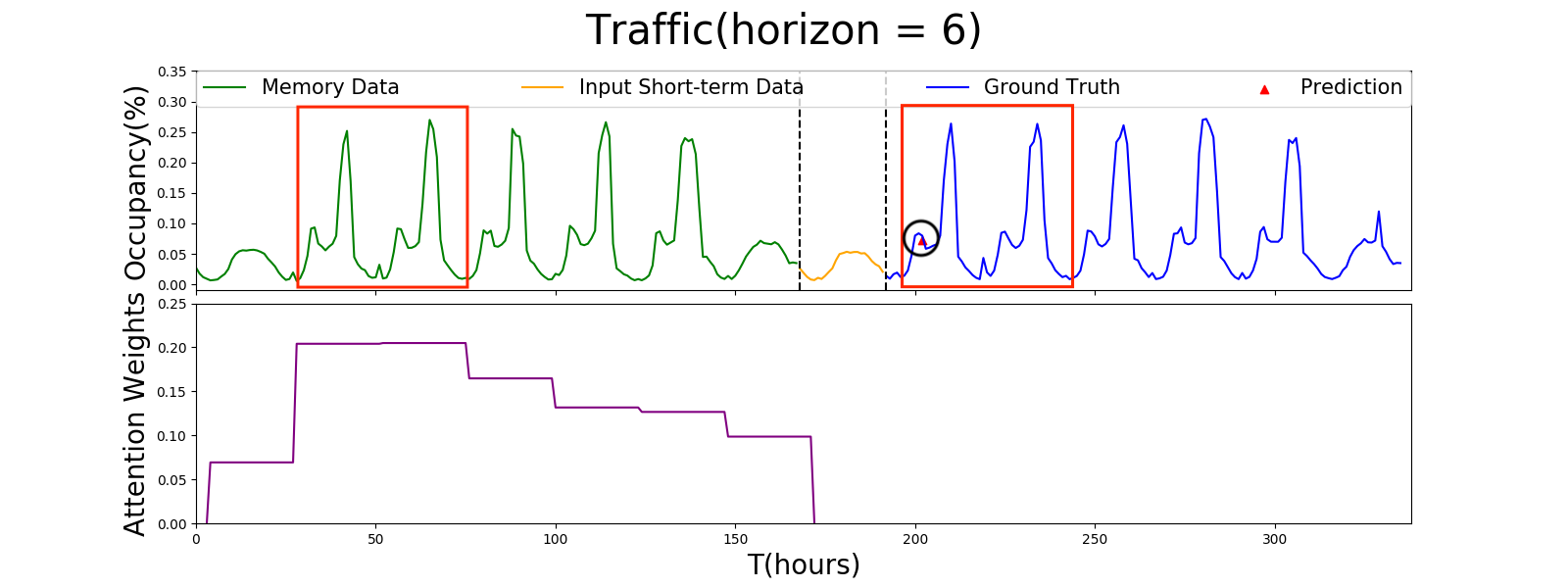}
   \caption{Plot of the attention weights between Input and Memory component for MTNet.}
   \label{heatmap}
\end{figure}

A nice property of MTNet is that its attention mechanism can be visualized to show which segment in the history contributes to the final forecasting. To demonstrate it, we randomly select a couple timestamp in the Traffic Occupancy Dataset to be forecasted and plot the attention weights between the memory component and input short-term data component particularly for this chosen timestamp. 
The results are presented in Figure \ref{heatmap}. The top two subfigures are the results for horizon 24 and the bottom two subfigures are the results for horizon 6. The input short-term data contains the latest 24 hours of historic data, which is presented in yellow between the two vertical black dashed line.
The long-term data in the memory component contains historic data of the past 7 days which is represented by the green line. The blue line denotes the time series ground truth. Purple curves below are the corresponding attention weights over all blocks in the memory at the chosen timestamp. The particular timestamp to be predicted is circled with black. Blocks with the highest weight and the block of data around the ground truth to be predicted are framed in red. We can observe that the attention mechanism automatically assign larger weights to the historical blocks which have highly correlated patterns with the ground truth. We can see that MTNet successfully learns to focus on corresponding blocks in the memory component to support the prediction.

\section{Conclusion}
In this paper, we proposed a memory time-series neural network (MTNet) inspired by end-to-end memory network. The model consists of a large memory component, three different embedding feature maps generated by three different encoders. This architecture can capture long-term dependencies while incorporating multiple variable dimension. Extensive experiments are conducted on 6 datasets that exhibit short-term and long-term repeating patterns and it demonstrated that our models can outperform state-of-the-art methods in both univariate and multivariate time series prediction. Moreover, block-wise attention over large memory component equips MTNet with great interpretability.

The future works include finding a better way to deal with rare events which now requires large amount of memory to be detected as well as improving the interpretability of the model. 

\vspace{-1cm}

\newpage
\balance
\bibliography{formatting-instructions-latex-2019}
\bibliographystyle{aaai}
\end{document}